\documentclass[sigconf]{acmart}
\usepackage{hyperref}

\AtBeginDocument{%
  \providecommand\BibTeX{{%
    \normalfont B\kern-0.5em{\scshape i\kern-0.25em b}\kern-0.8em\TeX}}}

\copyrightyear{2022} 
\acmYear{2022} 
\setcopyright{othergov}\acmConference[WWW '22 Companion]{Companion Proceedings of the Web Conference 2022}{April 25--29, 2022}{Virtual Event, Lyon, France}
\acmBooktitle{Companion Proceedings of the Web Conference 2022 (WWW '22 Companion), April 25--29, 2022, Virtual Event, Lyon, France}
\acmPrice{15.00}
\acmDOI{10.1145/3487553.3524250}
\acmISBN{978-1-4503-9130-6/22/04}

\begin{document}

\title{QAnswer: Towards Question Answering Search over Websites}

\author{Kunpeng GUO}
\authornote{Corresponding Author}
\orcid{0000-0002-0692-0057}
\affiliation{%
\institution{The QA Company SAS}
  \institution{Laboratoire Hubert Curien, UMR CNRS 5516}
  \country{Saint-Etienne, France}
}
\email{kunpeng.guo@univ-st-etienne.fr}

\author{Clement Defretiere}
\orcid{0000-0002-4471-2080}
\affiliation{%
\institution{The QA Company SAS}
\institution{Université Jean Monnet}
  \country{Saint-Etienne, France}
}
\email{clement.defretiere@the-qa-company.com}

\author{Dennis Diefenbach}
\orcid{0000-0002-0046-2219}
\affiliation{%
\institution{The QA Company SAS}
\institution{Laboratoire Hubert Curien, UMR CNRS 5516}
  \country{Saint-Etienne, France}
}
\email{dennis.diefenbach@the-qa-company.com}

\author{Christophe Gravier}
\orcid{0000-0001-8586-6302}
\affiliation{%
  \institution{Laboratoire Hubert Curien, UMR CNRS 5516}
  \country{Saint-Etienne, France}
}

\email{christophe.gravier@univ-st-etienne.fr}

\author{Antoine Gourru}
\orcid{0000-0003-3571-2430}
\affiliation{%
  \institution{Laboratoire Hubert Curien, UMR CNRS 5516}
  \country{Saint-Etienne, France}
}
\email{antoine.gourru@gmail.com}

\renewcommand{\shortauthors}{Guo, et al.}

\begin{abstract}
Question Answering (QA) is increasingly used by search engines to provide results to their end-users, yet very few websites currently use QA technologies for their search functionality. To illustrate the potential of QA technologies for the website search practitioner, 
we demonstrate web searches that combine QA over knowledge graphs and QA over free text -- each being usually tackled separately. We also discuss the different benefits and drawbacks of both approaches for  web site searches.
We use the case studies made of websites hosted by the Wikimedia Foundation (namely Wikipedia and Wikidata). 
Differently from a search engine (e.g. Google, Bing, etc), the data are indexed \textit{integrally}, i.e. we do not index only a subset, and they are indexed \textit{exclusively}, i.e. we index only data available on the corresponding website.
\end{abstract}

\ccsdesc[500]{Information systems~Web searching and information discovery}
\ccsdesc[500]{Information systems~Question answering}

\keywords{Website Search, Question Answering, Open Domain Question Answering, Knowledge Graph Question Answering}

\maketitle

\section{Introduction}

Search is a fundamental element of the web which is offered mainly on two levels:
\begin{itemize}
    \item Global Web search over (potentially) all websites available on the web, which is offered by search engines like (Google, Bing and Yandex)
    \item Search on a single Website scale in order to create a search experience particularly adapted to the given Website content.
\end{itemize}
Traditionally, both search functionalities exploit Information Retrieval (IR) systems. These are used to index the content and, given a search query, retrieve the documents that are more similar to the user's request. In the last years, more fine-grained search systems are emerging that are derived from Question Answering research~\cite{nakano2021webgpt}. These not only retrieve a document that potentially contains the answer, but also highlight (for extractive QA~\cite{joshi2020spanbert}) or structure (for Knowledge Graphs QA~\cite{diefenbach2018core}) the answer itself taken from the retrieved document which most likely contains the answer. These techniques are becoming more frequent in search engines that increasingly present ``direct answers'' to the user's queries. We believe that this trend will also move to website search with the aim of improving content access and user experience.\\
Websites hosted by the Wikimedia foundation are the most accessible websites to showcase search using question answering technologies. The main reason is that many researchers exploit these data due to the availability, licensing and quality of the data. As a consequence many resources (in terms  of approaches and training data) are available. In this demo paper, we present a website search based on question answering technologies. We index the content of Wikipedia and Wikidata~\cite{vrandevcic2014wikidata}. Both contents are currently only accessible on the respective websites via the traditional full-text search. The content includes both unstructured and structured contents. We therefore combine question answering technologies over free text and over knowledge graphs. We show the search experience that these technologies allow and describe the advantages that each one brings.

\section{Related Work}\label{sec:related}
Question Answering became a very active field of research in the last decade~\cite{diefenbach2018core, zhu2021retrieving} thanks to the progresses of neural networks that allow casting QA as different neural architectures. One can distinguish two main areas based on the type of underlying data: question-answering over free text (or open domain question answering) and question-answering over knowledge graphs.\\
\textbf{Question answering over free text} is generally tackled by marking the relevant answer of a user's question in a paragraph -- in this case, an annotated sample contains both the question, a paragraph, and the start as well as the end tokens of the answer span in the given paragraph. This is usually referred to as the task of Machine Comprehension. This field gained momentum with the publication of the Stanford Question Answering Dataset (SQuAD) \cite{rajpurkar2016squad}, a question-answering dataset that contains around 100k triplets (question, answer, and context). 
However, Machine Comprehension assumes that the paragraph containing the answer is given. In practice, such models (referred as \texttt{reader} models) must be coupled with an information retrieval system (referred as \texttt{retriever} models). One of the first retriever/reader architectures for open-domain QA systems is DrQA\cite{chen2017reading}. Another popular neural question-answering system is BERTSerini\cite{2019} that combines BERT \cite{devlin2019bert} with Anserini\footnote{\url{https://github.com/castorini/anserini}} (an information retrieval framework based on BM25). \\
\textbf{Question Answering over Knowledge Graphs (KGs)} is generally tackled by converting a user's question into a structured request (often a SPARQL query) over a knowledge graph~\cite{chakraborty2019introduction}. The result of the structured request is then shown as an answer to the user. One can distinguish two main lines of works. The first tries to construct possible structure requests using some heuristics and rank them to find the correct ones~\cite{diefenbach2020towards}. Other works try to solve the problem using generative models by translating the natural language request to the corresponding structured one~\cite{guo2018dialog}.\\ 
In this demo, we combine both technologies to offer a search over both unstructured and structured contents. We are not aware of any scientific publication describing this combination. Note that this differs from Hybrid search (like ~\cite{xu2016hybrid}) that tries to combine textual and knowledge graphs data to reply to a single question. In this work, we use the two sources for the search and provide the more relevant answer to the user.

\section{Description}

To showcase how current QA technologies can be applied on a website corpus we provide an online demo that, given a question, queries both Wikipedia and Wikidata -- the two wiki websites from the Wikimedia foundation that receive the most edits. %
We indexed Wikipedia's official released dump from the 1st January 2022 with a size of 19 gigabytes and Wikidata's release from the 22nd January 2022 with a size of 851 gigabytes in N-triples. Note that Wikipedia is a free text corpus while Wikidata is a structured dataset that can be downloaded as an RDF KG. 
Note that our design decisions are not only driven by best-performing approaches, but also by feasibility and scalability (in terms of data size and speed) constrains.\\
QA over free text part is supplied by a retriever/reader architecture~\cite{chen2017reading,2019,khattab2020colbert} as discussed in Section~\ref{sec:related}. %
For the retriever component, we choose BM25 with the default parameters in Anserini. We limit the number of returned documents to 29 to ensure the time efficiency of the pipeline. Furthermore, we split the original wikipedia articles into paragraphs which can not only improve the retriever accuracy~\cite{2019} but also boosts the inference process for the reader.
For the reader component, we choose RoBERTa~\cite{liu2019roberta} base model (12-layer, 768-hidden, 12-heads, 125M parameters) as pre-trained language model (PLM). RoBERTa inherits the architecture from BERT model, optimizes the key hyper-parameters, pre-trains the model in mini-batches, and removes the abundant training objective next sentence prediction which is proven to be insufficient for LM pretraining. As a result, RoBERTa outperforms BERT in many benchmarks like GLUE~\cite{wang2019glue}, SQuAD, and RACE~\cite{lai2017race}. To build our QA model, we apply two softmax layers to predict the start and end positions of the answer in the provided context following previous works~\cite{devlin2019bert}. Furthermore, we fine-tune the PLM using SQuADv2.0 which contains answerable and unanswerable questions. The unanswerable questions allow training the model to not answer which is essential in this case.\\\\
For QA over KGs, we choose the approach described in~\cite{diefenbach2020towards,diefenbach2019qanswer}. This approach allows to scale up to the size of the Wikidata KG with reasonable hardware and offers support for both keyword and natural language questions. The approach can be summarized as follows:
\begin{itemize}
    \item In a first step, one identifies possible nodes and relation in the graph the question could refer to. This is done by matching all possible n-grams in the question to nodes and relations in the graph having the exact same label up to stemming.
    \item In a second step, all previously identified nodes and relations are used to construct possible SPARQL queries. The approach is trying to construct only SPARQL queries that have non-empty results over that graph. This is achieved by looking at how the identified nodes and relations lie with respect to each other in the graph. This is done by performing a breadth-search starting from all nodes identified in the first step.
    \item The list of generated SPARQL queries is ranked based on different features in a \emph{learning to rank} manner. One of the features includes for example the number of words in the question that match the labels of the nodes used in a SPARQL query.
    \item Finally a confidence score for the first generated answer is computed using a binary classifier. 
\end{itemize}

To combine the two approaches, we employ a fallback pipeline: we first try to find an answer in the Wikidata KG. If the confidence is below the answer threshold we try to find an answer over Wikipedia. When both methods cannot provide an answer with high confidence that passes the threshold then we do not reply to the question (user can still choose to check the low-score answers).

\begin{figure}
  \href{https://wikimedia.qanswer.ai/qa/full?question=Scientific+conference+series+about+the+web\%3F&lang=en&kb=wikidata\%2Cwikipedia&user=open}{\includegraphics[width=0.5\textwidth]{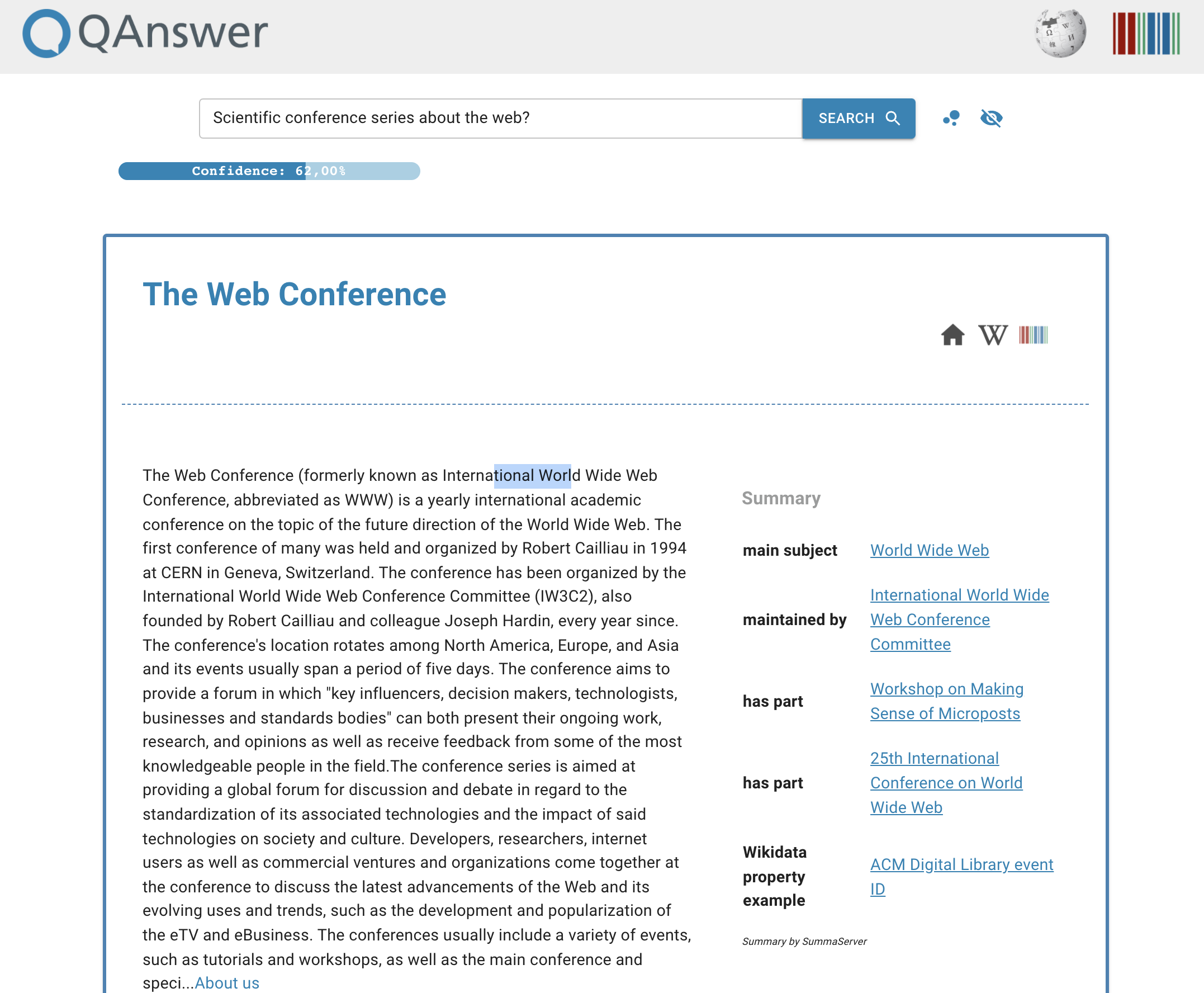}}
  \caption{Result of the query: "scientific conference series about the web". The screenshot show the answer together with contextual information like the description, the home page, the Wikipedia and Wikidata link, an image and a summary.}
  \label{fig:panel}
\end{figure}

In the following, we want to describe and illustrate some of the main advantages of each of the two QA approaches. In general, QA over KGs is a technology of choice to query structured data (since structured knowledge can be easily represented in a KG). On the other hand, open domain question answering is a technology of choice to query a document collection containing unstructured data which has not been encoded in a KG.\\

\begin{figure}
\href{https://wikimedia.qanswer.ai/qa/full?question=who+participated+in+the+web+conference+2018&lang=en&kb=wikidata\%2Cwikipedia&user=open}{\includegraphics[width=0.5\textwidth]{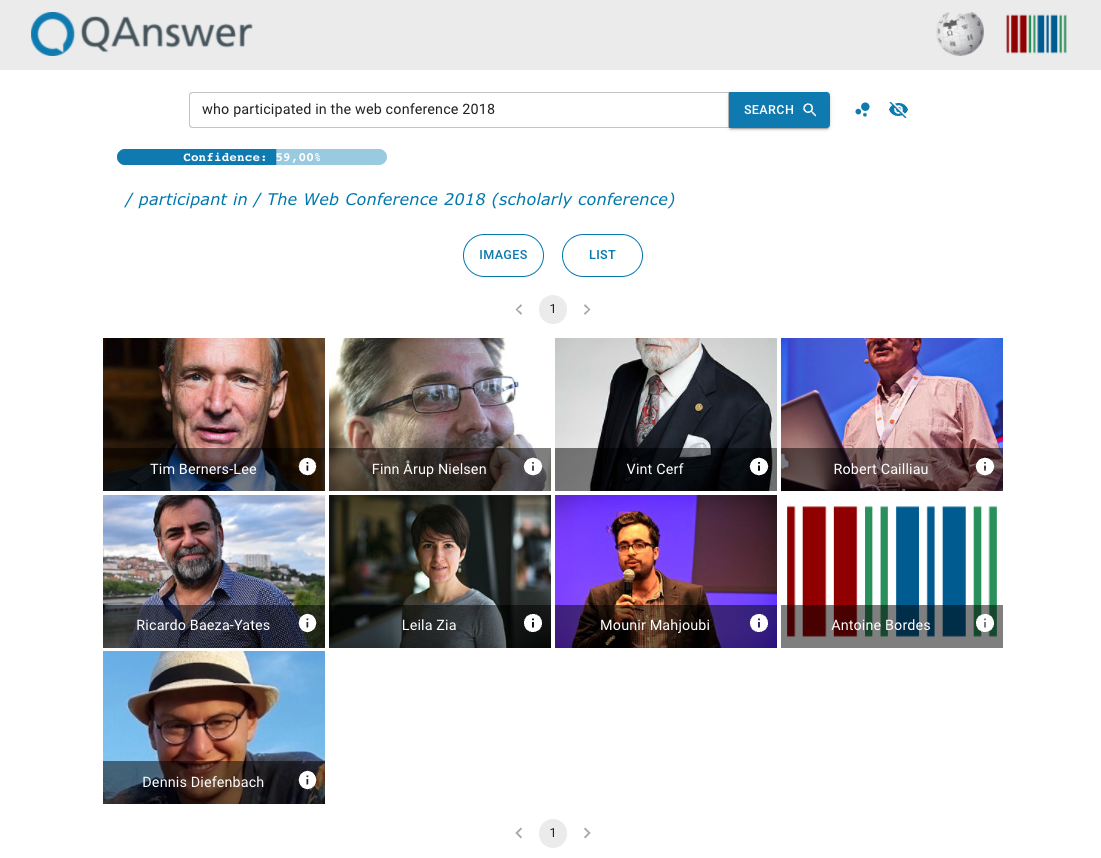}}
  \caption{Result of the query: "who participated in the web conference 2018". The screenshot shows the answers aggregated in a grid list. When the answer set contains multiple answer entities with metadata (like images or geo-coordinates), we aggregate these information in a grid, or map view.}
  \label{fig:aggregation}
\end{figure}

Question Answering over KGs has mainly two benefits over Question Answering over Free Text. The first is that the data in KGs can not only be used to retrieve the answer but can also be used to show contextual information about the answer itself. This includes external links, images, maps, descriptions, and summaries. This is for example shown in Figure~\ref{fig:panel}, where we show the answer for the query "scientific conference series about the web". This is not doable with question answering over text, where only the answer string itself can be shown. Some other examples showcase the contextual information\footnote{Please note that all questions here are also actual links to the demo system online}: 
\begin{itemize}
    \item \href{https://wikipedia.qanswer.ai/qa/full?question=What\%27s+the+capital+of+Italy\%3F&lang=en&kb=wikidata\%2Cwikipedia&user=open}{What's the capital of Italy?}
    \item \href{https://wikipedia.qanswer.ai/qa/full?question=Who+is+the+current+UK+prime+minister\%3F&lang=en&kb=wikidata\%2Cwikipedia&user=open}{Who is the current UK prime minister?}
\end{itemize}
Another advantage is that it is easier to aggregate information. When the answer is represented by multiple entities, it is possible to aggregate the contextual information into image grids and maps. We demonstrate this in Figure \ref{fig:aggregation}, with the answer to the query "who participated in the web conference 2018". In this case, the answer is represented by multiple entities and the answer images are aggregated into a single image grid. This is not possible for question answering over free text where the results are generally distributed across many documents. We will showcase to the attendees other examples demonstrating the aggregation feature such as:
\begin{itemize}
    \item \href{https://wikipedia.qanswer.ai/qa/full?question=show+me+Museums+in+Rome&lang=en&kb=wikidata\%2Cwikipedia&user=open}{Show me museums in Rome?}
    \item \href{https://wikipedia.qanswer.ai/qa/full?question=Cinemas+in+London\%3F&lang=en&kb=wikidata\%2Cwikipedia&user=open}{Cinemas in London?}
\end{itemize}

\begin{figure}
\href{https://wikimedia.qanswer.ai/qa/full?question=Where+is+the+web+conference+taking+place&lang=en&kb=wikidata\%2Cwikipedia&user=open}{\includegraphics[width=0.5\textwidth]{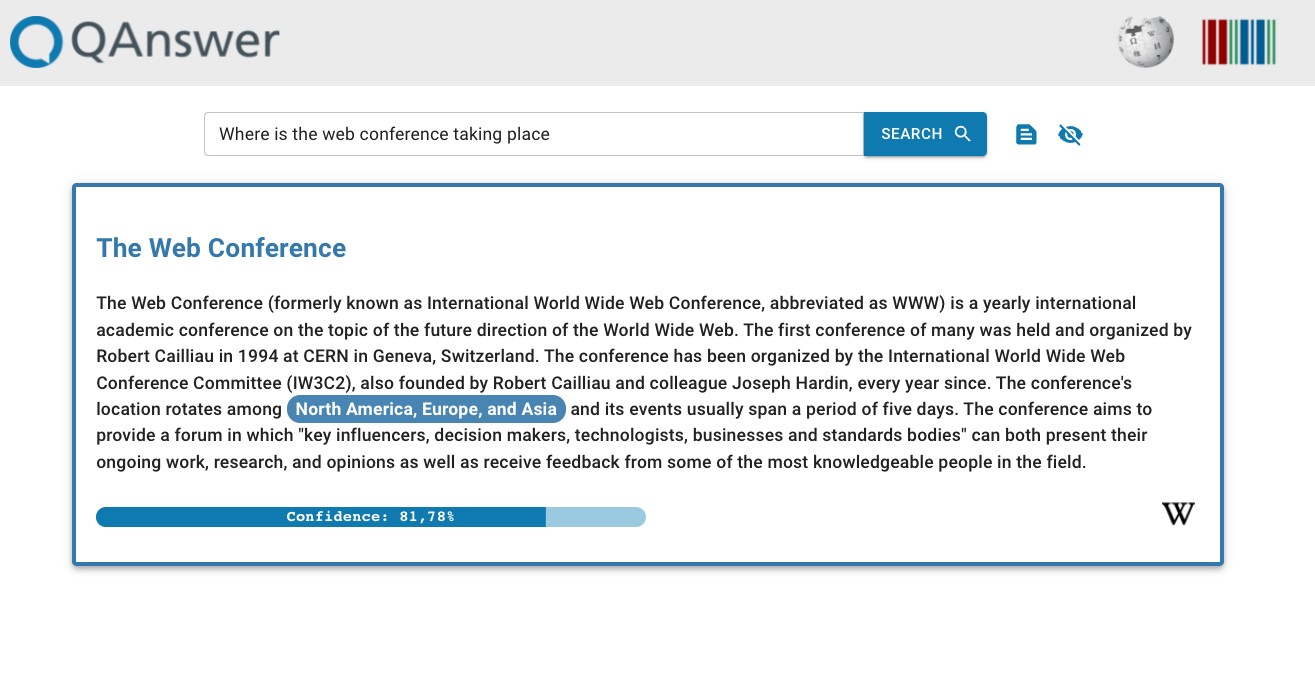}}
  \caption{Result of the query: "Where is the web conference taking place". The screenshot shows the answer marked in the corresponding paragraph where it was found. A link points to the corresponding paragraph in Wikipedia. For browsers like Chrome, the link will not only point to the Wikipedia page, but will also mark the retrieved answer \emph{in the page}.}
  \label{fig:text}
\end{figure}

\begin{figure}
  \href{https://wikimedia.qanswer.ai/qa/full?question=reasons+to+attend+to+the+web+conference&lang=en&kb=wikidata\%2Cwikipedia&user=open}{\includegraphics[width=0.5\textwidth]{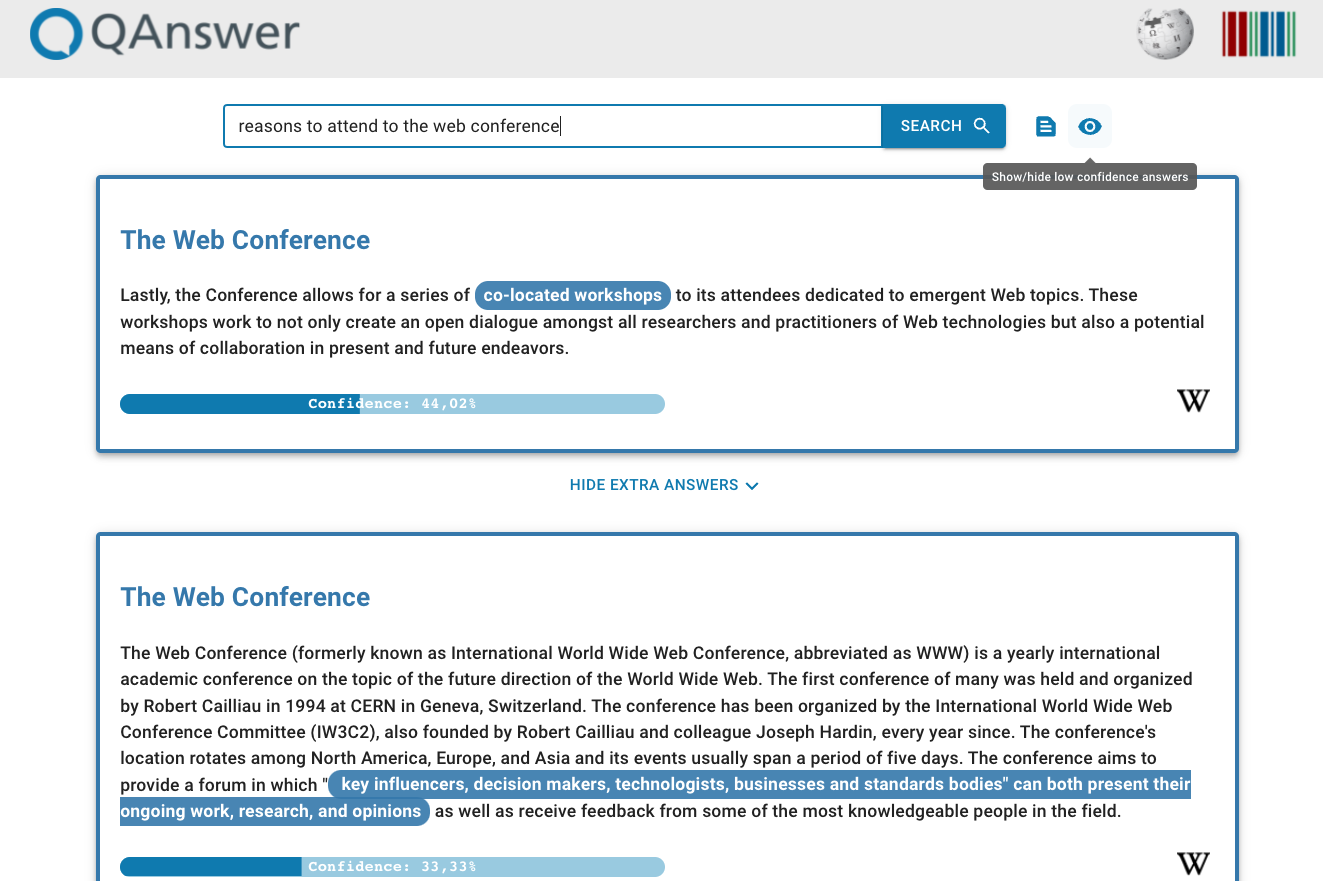}}
  \caption{Result of the query: "reasons to attend to the web conference". The screenshot shows several paragraphs containing possible answers. Note that these answers have low confidence and are only rendered when clicking on the eye-shape button on the page. This is due to the nature of the question, which is closer to an exploratory search and for which no ``explicit answer'' can be found.}
  \label{fig:multiple}
\end{figure}

Complementary, Question Answering over free text also has numerous advantages. One main advantage is that not only the answer is provided, but also an explanation is provided (context from which the answer is taken). The surrounding text in a paragraph is the reason why the language model selected the answer span as an answer. This context also explains the answer and can help the user to judge the result (see for example Figure~\ref{fig:text}). This is not possible over Knowledge Graphs where there is generally no explicit information of why this fact is contained in the graph. The query itself can be used to give an interpretation of what was understood, but not why the answer is true. Some other examples demonstrate the QA over free-text:
\begin{itemize}
    \item \href{https://wikipedia.qanswer.ai/qa/full?question=How+many+votes+Joe+Biden+got\%3F&lang=en&kb=wikidata\%2Cwikipedia&user=open}{How many votes Joe Biden got?}
    \item \href{https://wikipedia.qanswer.ai/qa/full?question=How+Spider-Man+got+his+abilities\%3F&lang=en&kb=wikidata\%2Cwikipedia&user=open}{How Spider-Man got his abilities?}
\end{itemize}
Another advantage is that for ambiguous questions like: "reasons to attend to the web conference" while there is no direct answer, QA models can still present insightful options (see Figure~\ref{fig:multiple}. This is not possible in question answering over KGs where such fine-grained meanings cannot be encoded. Some other examples showcase this advantage:
\begin{itemize}
    \item \href{https://wikipedia.qanswer.ai/qa/full?question=Why+people+cry\%3F&lang=en&kb=wikidata\%2Cwikipedia&user=open}{Why people cry?}
    \item \href{https://wikipedia.qanswer.ai/qa/full?question=What\%27s+the+causes+of+global+economic+crisis\%3F&lang=en&kb=wikidata\%2Cwikipedia&user=open}{What's the causes of global economic crisis?}
    \item \href{https://wikipedia.qanswer.ai/qa/full?question=How+should+protect+ourselves+from+COVID-19\%3F&lang=en&kb=wikidata\%2Cwikipedia&user=open}{How should protect ourselves from COVID-19?}
\end{itemize}

Note that differently from a search engine, we indexed the underlying website content integrally. Search engines generally only index subsets of the dataset and exploit it for question answering. For example, the question "who participated in the web conference 2018" in Figure~\ref{fig:multiple} will not give a ``direct answer'' in google search probably because, for these types of information, the Wikidata KG is not seen as complete enough. Additionally, results to exploratory questions like "reasons to attend to the web conference" (Figure \ref{fig:multiple}) are not in the scope of ``direct answers'' in a search engine. Moreover, we index the website content exclusively. A search engine might choose other relevant websites to find the answer to a user's question. But this might not be the goal or the interest of the website curator.

\section{Demo}
A demo of the current version can be found under:

\centerline{\href{http://wikimedia.qanswer.ai}{http://wikimedia.qanswer.ai}}

Moreover, by clicking on the above figures or examples you will be redirected to the online demo.

\section{Conclusion}
With this demo, we demonstrate how current question answering technologies can be used for providing search on website content. We provide question answering search over both the unstructured content of Wikipedia and the structured content of Wikidata. We achieve this by combining an open domain question answering system and a question answering system over KGs. The chosen methods are often used as baselines in current research publications but hardly with a website content corpus in mind. The goal is to give more awareness to the practitioner about the current status of question answering technologies, about their current capabilities, and about the opportunities they provide to shape website search in the future.

\bibliographystyle{ACM-Reference-Format}
\bibliography{main}

\end{document}